\documentclass{article}

\usepackage{PRIMEarxiv}

\usepackage[utf8]{inputenc} 
\usepackage[T1]{fontenc}    
\usepackage{hyperref}       
\usepackage{url}            
\usepackage{booktabs}       
\usepackage{amsfonts}       
\usepackage{nicefrac}       
\usepackage{microtype}      
\usepackage{lipsum}
\usepackage{fancyhdr}       
\usepackage{graphicx}       
\graphicspath{{media/}}     
\usepackage{color}
\usepackage{ifsym}
\usepackage{bbding}
\newcommand{\M}{\emph{MDsrv }}
\newcommand{\m}{\emph{MDsrv}}
\newcommand{\C}[1]{\begingroup\color{black} #1\endgroup}
\newcommand{\CN}[1]{\begingroup\color{black} #1\endgroup}

\title{MDsrv - visual sharing and analysis of molecular dynamics simulations}

\author{
 Michelle Kampfrath$^\S$\\
  Leipzig University\\ Image and Signal Processing Group\\ Leipzig, Germany \\
   \And
  René Staritzbichler$^\S$ \\
  Leipzig University\\ Institute of Biophysics and Medical Physics\\ Leipzig, Germany \\
     \And
  Guillermo Pérez Hernández \\
  Charité Universitätsmedizin Berlin\\ Institute of Medical Physics and Biophysics\\ Berlin, Germany \\
       \And
  Alexander S. Rose \\
  Mol* Consortium\\\hspace{6cm} \\
         \And
  Johanna K.S. Tiemann \\
  University of Copenhagen\\ Linderstrøm-Lang Centre for Protein Science\\  Copenhagen, Denmark\\
           \And
  Gerik Scheuermann \\
    Leipzig University\\ Image and Signal Processing Group\\ Leipzig, Germany\\ \hspace{8cm}\\
               \And
  Daniel Wiegreffe$^\S$\Envelope \\
    Leipzig University\\ Image and Signal Processing Group\\ Leipzig, Germany \\
                   \And
 Peter W. Hildebrand$^\S$\Envelope \\
   Leipzig University\\ Institute of Biophysics and Medical Physics\\ Leipzig, Germany\\\
}

\begin{document}
\maketitle

Correspondence: daniel@bioinf.uni-leipzig.de and peter.hildebrand@medizin.uni-leipzig.de\\
$^\S$ Authors contributed equally\\
\begin{abstract}
Molecular dynamics simulation is a proven technique for computing and visualizing the time-resolved motion of macromolecules at atomic resolution. The MDsrv is a tool that streams MD trajectories and displays them interactively in web browsers without requiring advanced skills, facilitating interactive exploration and collaborative visual analysis. We have now enhanced the MDsrv to further simplify the upload and sharing of MD trajectories and improve their online viewing and analysis. With the new instance, the MDsrv simplifies the creation of sessions, which allows the exchange of MD trajectories with preset representations and perspectives. An important innovation is that the MDsrv can now access and visualize trajectories from remote datasets, which greatly expands its applicability and use, as the data no longer needs to be accessible on a local server. In addition, initial analyses such as sequence or structure alignments, distance measurements, or RMSD calculations have been implemented, which optionally support visual analysis. Finally, \C{based on Mol*}, MDsrv now provides faster and more efficient visualization of even large trajectories \C{compared to its predecessor tool NGL}.

\end{abstract}

\section*{Graphical Abstract}
\begin{figure}[h]
\begin{center}
\includegraphics[width=1.0\columnwidth]{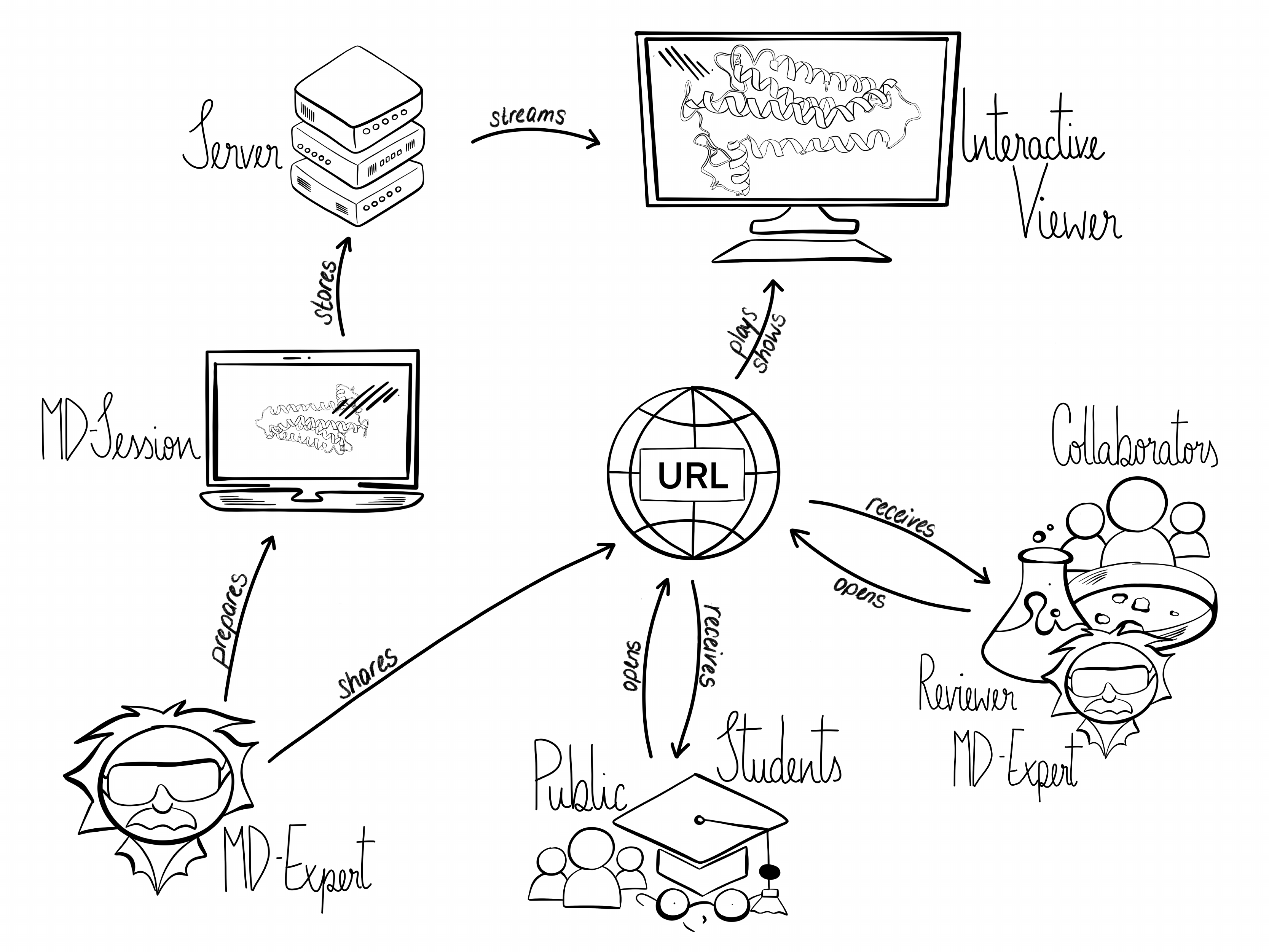}
\end{center}
\caption{\C{MDsrv helps experts prepare interactive MD sessions and share them by simply sending a URL link to students, staff, or by including the URL in publications.} }
\label{NAR-fig0}
\end{figure}

\section*{Introduction}
\enlargethispage{-65.1pt}

Molecular dynamics (MD) simulations are a well-established technique to investigate time-resolved motions of biological macromolecules at atomic resolution~\cite{dror2012biomolecular,perilla2015molecular}. MD simulations are routinely applied to answer many biologically relevant questions and have proven their value in deciphering functional mechanisms of proteins such as of G protein-coupled receptors (GPCRs), DNA, or RNA and thereby aiding also in the design of novel drugs (e.g.~\cite{suomivuori2020molecular}). Recent approaches using machine learning that push the limitations of classical MD simulation~\cite{jia2020pushing} give a glimpse of future applications that may even exceed the time scales of coarse-grained dynamics \cite{yu2021multiscale,gupta2022ugly}. MD simulations can provide unprecedented insight into temporal and spatial details by calculating the dynamic behavior of proteins in their natural cellular environment in silico.

In recent years, researchers using MD simulations have helped make the field of MD more FAIR (findability, accessibility, interoperability, reproducibility of data~\cite{wilkinson2016fair}) by depositing their data in general-purpose scientific data repositories, making them accessible for further study and research~\cite{abraham2019sharing}. Even if access to MD trajectories is granted, some degree of expert knowledge is required to even load and visually examine a simulation. Visualization itself is a great way to bridge technical knowledge gaps, for example between experimental and computational researchers, as it facilitates and guides data analysis and the complementary strengths of human and machine analysis are potentiated when led by interactive visualization~\cite{crouser2016toward}. Web-enabled visualization solutions help reach a wider audience. Web visualization tools are now increasingly being developed and made available for visualizing MD trajectories. \C{The popularity of these applications is also due to the free and easy accessibility of the tools for visualization of MD trajectories and their user-friendliness.}

The \M was the first tool that made interactive streaming and visualization of MD trajectories in web browsers more accessible~\cite{Tiemann2017}; others followed including HTMoL~\cite{carrillo2018htmol}, UnityMol~\cite{martinez2021unitymol}, PCAViz~\cite{pacheco2019pcaviz}, and 3dRS~\cite{bayarri20213drs}. \C{The applications of these tools are broad, and they all cover the same functionalities of in-browser visualization and sharing of remote data. Each one, however, differentiates itself by adding  different features, e.g. the programmatic manipulation of the trajectory via text-console~\cite{carrillo2018htmol}, the visualization of principal components~\cite{pacheco2019pcaviz} or of arbitrary collective variables~\cite{bayarri20213drs}. As of to our knowledge, with its simplicity, the \M and in particular the NGL Viewer allowed an easy and streamline embedding into other websites. An example is the GPCRmd platform,} which hosts MD simulations of more than 200 GPCRs and visualizes the intrinsic flexibility of the 3D-GPCRome that in turn provides valuable insights into signaling processes that are also relevant to drug development~\cite{rodriguez2020gpcrmd}. \C{Furthermore, the NGL Viewer and \M were the basis for 3dRS that improved even more the user experience and facilitate session generation with the goal to make such MD streaming viewers more applicable by the community. 3dRS is well adapted to sharing structures and trajectories, but only up to a certain size. Due to the ever increasing amount of data, MDsrv addresses this aspect by being optimized for arbitrarily large data. This approach led in the development also to smaller restrictions (e.g., possible file formats) that will be addressed in future releases.}

\M already allowed MD experts to set up a session to share an intuitive visualization of a trajectory with collaborators. Although it has been adopted by many computational biophysicists, usability was still somewhat limited. First, creating a session required knowledge of JavaScript, and extracting accurate representations, colors, or alignments was tedious. Simple and helpful analysis functions such as distance measurements required even more expert knowledge of the underlying functions of the \M which in turn relied on the NGL Viewer visualization~\cite{Rose2015}. This complicated the application and limited the visualization setup to experts. 
Here we introduce a new instance of \M that allows non-experts to load and share MD trajectory sessions more easily (\C{see as overview Fig \ref{NAR-fig0}, available at }\url{https://proteinformatics.informatik.uni-leipzig.de/mdsrv}).

The underlying viewer, NGL, has been replaced by its successor Mol*~\cite{Rose2021}, which combines the strengths of the NGL Viewer and the LiteMol Suite~\cite{sehnal2017litemol}. Mol* offers a number of benefits for \m, including superior visual quality, improved secondary structure assignment, trajectory animation video export, and fully GUI-driven session support. We created a custom viewer built on top of Mol*. \C{This approach not only allowed the \M to directly benefit from the above-mentioned features of Mol*, but also opened up the possibility of integrating necessary changes directly into Mol*, in addition to creating further specifications at the level of the MDsrv.}  
 We contributed an importer for Zenodo (\url{https://zenodo.org/}) datasets to Mol* that greatly simplifies access and visualization of trajectories from Zenodo and alleviates the need for one's own \M instance. \C{In addition, interactive trajectory analysis features for distance, RMSD, angle, or dihedral measurements have been added. Finally, we modularized the software for MDsrv in containers to simplify the installation of MDsrv. With our update of the \m, we take another step towards our goal of making MD simulations even more accessible to a broader user base in line with the FAIR principles.}

\section*{Methods and Description}

\subsection*{Server}

\begin{figure*}[t]
\begin{center}
\includegraphics[width=0.9\textwidth]{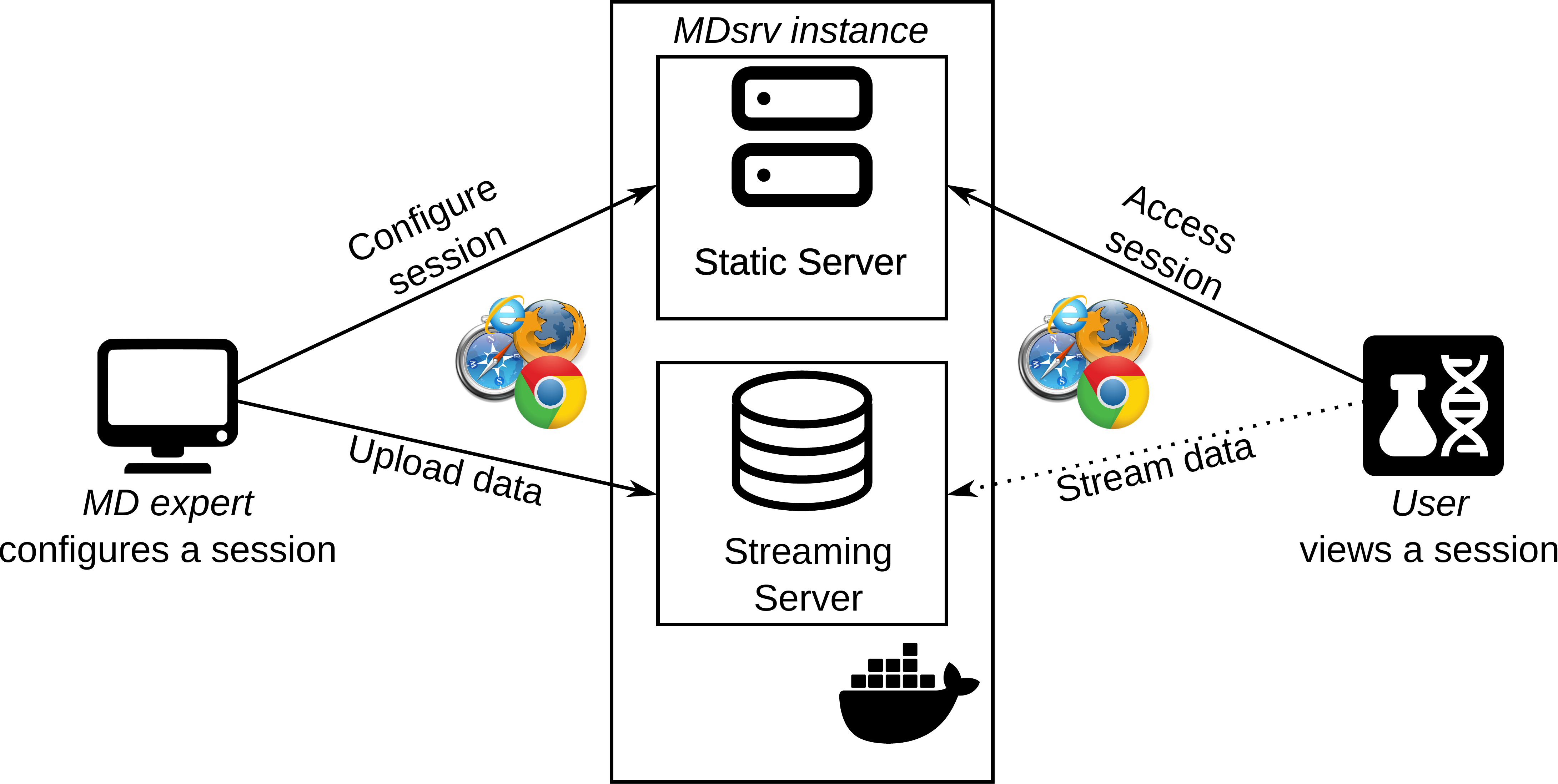}
\end{center}
\caption{The system overview of \m. MD experts can configure a new session via the frontend \C{provided by the static server} of \M and store the data on the Streaming Server of \m. Afterward, other users can access this session and stream the provided MD simulation. The streaming works automatically without any additional user interaction.
}
\label{NAR-fig1}
\end{figure*}

\textit{General architecture:}
The \M web service is designed for sharing and interactively viewing of trajectories from MD simulations. It can be used, for example, to facilitate collaboration between researchers working in experimental and computational groups on different continents. To this end, we provide \m, a platform (see Fig. \ref{NAR-fig1}) where MD trajectories can be uploaded and their visualization configured. \M is divided into two components. The \C{static server} provides access to the \M website and delivers all the functionality of \M to the client.  
In addition, the streaming server delivers the MD trajectories frame by frame. The client's local web browser then processes this data locally in the \C{frontend} provided by the \C{static server}. This separation allows \M to scale easily since only the streaming server needs to be expanded in case of heavy usage. 
The MD expert, who delivers the trajectory data can in the next step design so-called sessions that are saved in the \M to make them publicly available. Such sessions are preconfigured visualizations e.g. a zoom on a ligand-binding pocket or a specific representation. For this purpose, the MD expert or any user can either directly upload their local trajectories or load publicly available data, such as those deposited at Zenodo directly into \m,  since \M can obtain them via their record number.
By accessing the frontend of \m, a user can then open these sessions and get the preconfigured visualization presented, where the view, zoom, and also other settings can still be manipulated interactively.

\textit{Streaming server:}
With the increase in the size of MD simulation trajectories, sharing and visualization becomes a challenge, especially when the trajectory must be loaded over the World Wide Web and then read in entirely by the visualization program. Meanwhile, the size of an MD trajectory can quickly exceed the size of a standard desktop computer's memory and in order to not crash the application, the transferred data of the trajectory must be reduced.  We, therefore, introduced a streaming server (see Fig. \ref{NAR-fig1}) for storing and processing before streaming the trajectory to the viewing client. Sharing MD trajectories via a web application has now been further enhanced for \m. 
For this purpose, a session is stored on the streaming server after being saved by the MD expert. This streaming server then sends only the necessary simulation steps to the user on request, which makes the memory consumption on the user side negligible. This allows, for example, to open a several gigabyte large MD simulation on a mobile device.

\textit{Docker container virtualization:}
We have modularized the software for \M in containers, as this allows us to simplify the installation of the web service, to offer the software in a more sustainable way, and additionally to provide an easy local installation. 
In order to make the sessions reusable according to the FAIR principles, \M should also be sustainable in terms of installation and reusability. To archive a session permanently, you need not only the data but also an instance of \M that was used to process the data, since future updates might change the results, e.g. the representation or the distance plots, depending on their default criteria.
The software must also be stored in a platform-independent manner to simplify future execution. A promising approach here is the container virtualization by Docker. We have wrapped the frontend and the streaming server of \M (see Fig~\ref{NAR-fig1}) in containers, so that \M can be operated independently of the underlying operating system. This makes it possible to deploy \M in the future on systems that would no longer be able to do so due to software updates. This principle is also successfully implemented by, for example, the Galaxy Browser~\cite{gruning2018practical}. 
For general usage without the necessity to install \M locally, as it was previously the case, we provide a publicly available instance of \M where all sessions are publicly accessible (\url{https://proteininformatics.informatik.uni-leipzig.de/}). To allow MD experts to share private MD trajectories e.g. with a reviewer during review process, \M can be easily installed on dedicated hardware by using the Docker images of the frontend and the streaming server.

\textit{Web viewer:}
We updated the underlying viewer application in \M from NGL to Mol*. A crucial feature of Mol* is its support for saving the application state at any time to reload it later. This allows \M to offer the creation of sessions without the need for programming knowledge. \C{MDsrv extends this functionality so that the previously local sessions can now be persistently stored using MDsrv in a publicly accessible manner along with the associated data.} To fulfill all the needs of \M we made a number of contributions to Mol*. 
We added support for more coordinates and topology file formats \C{that can be accessed from the client locally}, namely, top, prmtop, trr, and nctraj, in addition to the existing support for psf, dcd, and xtc. Further, we added a Zenodo import extension that allows easy loading of structures, trajectories, and zip archives from any Zenodo dataset. Finally, we leveraged the modular GUI of Mol* to create a custom viewer for \M that adds GUI components for MD analyses including linked time-traces of measurements and precomputed sequence alignments.

\C{
\textit{Limitations:}
The software architecture we have chosen has limitations in addition to the aforementioned advantages. Although \M has only low requirements on the client's hardware, the streaming server must have sufficient computing and storage capacity to process the trajectories. Furthermore, for a public usage scenario with multiple users, the server should be connected to the internet with sufficient network bandwidth. In addition, the streaming server could be hosted in a content delivery network, so that it is optimally accessible worldwide. 
For users to be able to stream the trajectories, they need a stable internet connection. 
}

\subsection*{\C{MD analysis}}
The analysis of MD trajectories can range from simple visual inspection of salient features, like (un)binding events, domain motion, interface formation etc, to more data-driven approaches that uncover relationships and events that might otherwise go unnoticed, like subtle bond-breaking, allosteric effects, and transient effects. 
In order to integrate the visual with the data-driven analysis, some molecular visualization tools (e.g., VMD~\cite{HUMP96} or PyMOL\footnote{Schrödinger, LLC (November, 2015) The PyMOL Molecular Graphics
System, Version 1.8.}) offer the on-the-fly computation of a variety of (geometric) measurements on the displayed geometry. 
This way, time-traces of distances, angles, dihedrals and RMSDs can be visualized \textit{in-program}, without resorting to a growing number specialized standalone libraries (MDanalysis~\cite{gowers2019mdanalysis,michaud2011mdanalysis}, MDtraj~\cite{McGibbon2015MDTraj}, MDciao\footnote{Pérez Hernández, G. and Tiwari, S. gph82/mdciao: https://zenodo.org/record/5643177. (November, 2021)}, BioPython~\cite{chapman2000biopython}). 

On par with the visualization programs mentioned above, \m~can compute time-traces of the geometric parameters. However, \m~goes one step further in user-friendliness, by making these time-traces interactive and navigable per mouse-click. Key events that might go unnoticed in the visual inspection, become sharp transitions in the navigable time-trace (see Fig. \ref{NAR-fig2}). Just by clicking on these transitions (or anywhere in the time-trace), the matching 3D frame will also appear in the main panel, making it easier to correlate events in the time-trace with events in the 3D trajectory. When combined with the sharing capabilities of \m, these navigable time-traces make the analysis considerably more accessible to the un-trained user, who might otherwise not distinguish these transitions from the continuous oscillations inherent to MD. Additionally, the values of the plot can be sorted or filtered to a given numerical range (e.g. when a distance between two atoms is between 3 and 4 $\AA$), making geometries/frames of interest even easier to locate.

\textit{Sequence Alignment} is one the most widely used tools in bioinformatics for a broad range of purposes~\cite{Stamm2013,Khafizov2010}. 
Alignments provide insight into both similarities as well as differences between pairs or multiple sequences. 
The full interpretation of an alignment, however, requires mapping the alignment onto 3D structures. 
Only in a 3D context can the spatial relationships between distant sequence positions become apparent. 
\M now offers as a novel feature the upload of an alignment together with two or more structures. By uploading multiple structures and an alignment, one can overlay or superimpose the structures based on the alignment.

\subsection*{Usage}

Creating, preparing and sharing sessions can be done via our provided server, or you can set up your own using Docker. 

\textit{Sharing a prepared session using a URL:}  
A new session can be created by selecting \textit{Start new Session} on our server (\url{https://proteininformatics.informatik.uni-leipzig.de/mdsrv}), or by setting up your own.
Using the Home menu on the left, structures and trajectories can be uploaded directly into the client from remote data repositories or from local sources.
Afterward, the session can be prepared as desired and shared via a URL. 
The \textit{Remote Session} feature in the lower panel is used for this purpose. 
To further describe the session, it can be given a name\CN{, description and source to document the origin of the displayed data and establish authorship for the session. This information is then displayed when such a session is used.}
After selecting the \textit{Upload} button, the session will be stored on our server.
If the session is to be stored on another \M instance, the address of the target server can be changed \CN{in the webinterface}. 
The uploaded session will now appear in the list of sessions in the \textit{Remote Session} \CN{list on the left side in the home} menu. 
Right-clicking on the session opens it in a new tab with its own URL. 
This URL can then be shared. 
An already saved session can also be further modified and uploaded again as a new session.

\textit{Upload and stream a trajectory:}
The \textit{Add Trajectory to Stream Server} feature enables the user to upload a trajectory to our streaming server instead of directly viewing the simulations in our frontend which would normally be too large to be held in the memory provided for the visualization client on the users' machine. 
Once the URL to the file, a name,\CN{a more detailed description, and source} is provided, the file will be saved on the server \CN{and is made available to users when viewing this data.}
\C{The streaming server always downloads the data to be streamed using a URL, since the data to be streamed is usually not stored on client devices due to its size. 
Since the streaming server can also be deployed in local environments, users can use their local instance to host and stream non-public data in their access-restricted usage scenario (e.g., within a lab).}
The trajectory will now be available to stream in the \textit{Match Stream Trajectory} feature.
A corresponding structure that matches the trajectory must be imported into the client to be able to view the trajectory frame by frame.
The trajectory can be played once it has been matched to the structure using the \textit{Match Stream Trajectory} feature.
\C{Currently it is only possible to stream xtc formats, but we plan to expand the supported formats in the future.}

\textit{Time-traces for trajectories:}
We extended the measurement function for structures and trajectories by adding the possibility to display a time-trace of the measurement over all frames for a trajectory.
After importing a trajectory and adding a measurement by selection, e.g. 2 atoms/residues for the distance or 3 for an angle, its time-trace can be displayed in the \textit{Time-trace plot} feature. 
Different interactions with the plot are possible, e.g. sorting by frame, ascending, or descending, filters for values, selection of a specific value in the plot to skip to the corresponding frame, and displaying the RMSD value for the whole structure where the comparison frame can be defined.

\C{
\textit{Zenodo import:}
Files from a Zenodo record can be imported using the \textit{Zenodo Import} feature. 
The user only has to specify the \textit{Record} number to get the associated files displayed for import.
The \textit{Type} parameter allows to select what kind of data should be imported: structure, trajectory, volume, or compressed.
All the files to the corresponding type will be displayed in a drop-down menu below.
If there are no supported files in the record, the user will be notified.
}

\textit{Alignment:}
\C{Based on the functionality of Mol* to superpose structures, \M extends this functionality by additionally importing externally generated sequence alignments. The quality of a structural alignment depends crucially on the quality of the underlying sequence alignment. Obtaining high quality sequence alignments is far from trivial and often out of the scope of the default methods also implemented in Mol*. }
It is possible to superpose two or more structures based on an imported alignment. 
After a ClustalW file and additionally the structures from the alignment have been provided, the sequences in the alignment have to be matched with the corresponding sequences in the structures. 
The matching is done using the \textit{Match Sequence Alignment} feature. 
Once the alignment is complete, the superposition of the structures is calculated based on the alignment. 
The alignment can be displayed in the \textit{View Sequence Alignment} feature.

More detailed descriptions including video materials on how to use the features are available online (\url{https://proteininformatics.informatik.uni-leipzig.de/mdsrv/}).

\subsection*{Example Output}
\begin{figure*}[!t]
\begin{center}
\includegraphics[width=0.7\textwidth]{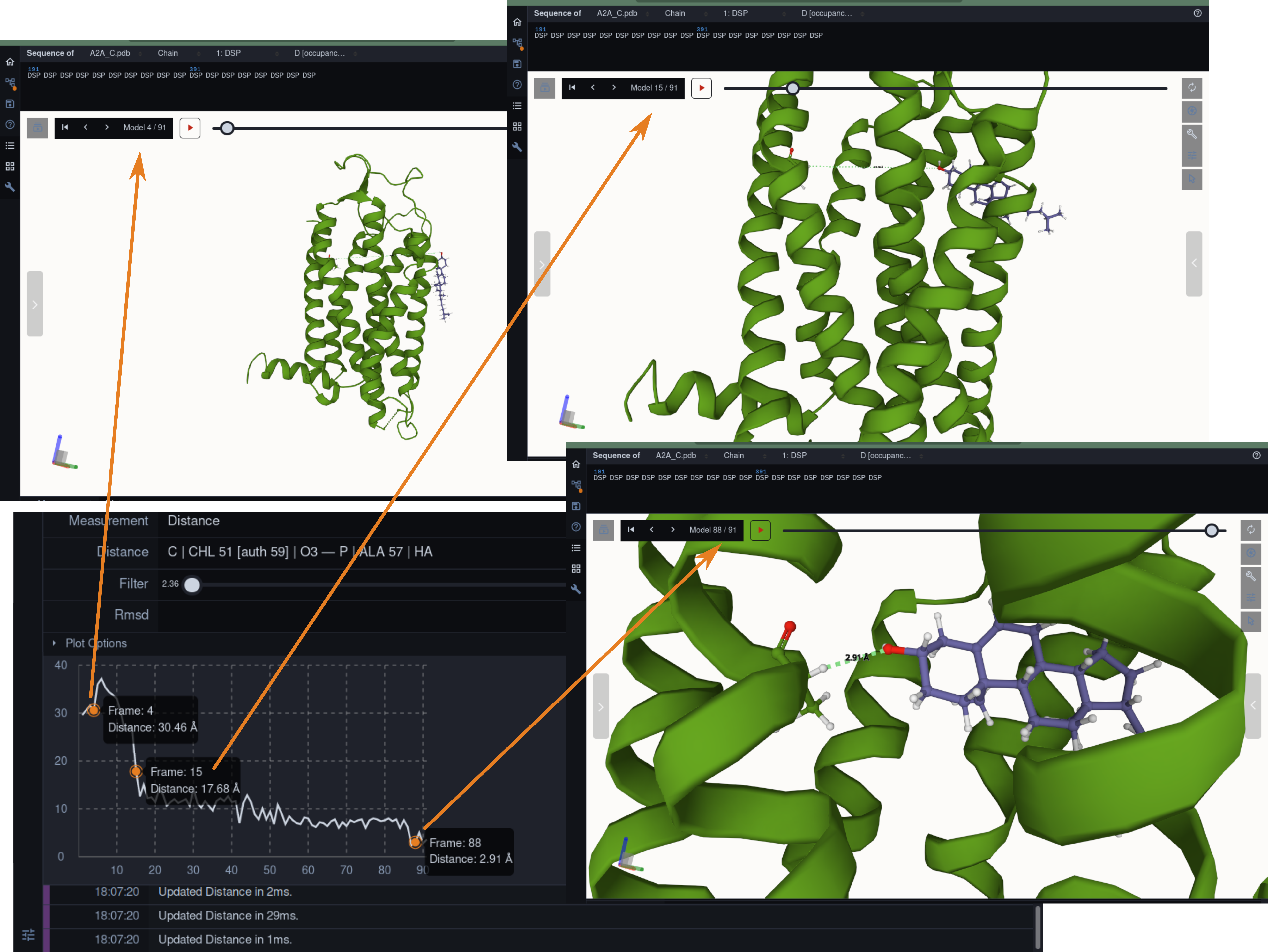}
\end{center}
\caption{Composite of screenshots showing the navigable time-trace (bottom-left) functionality. We have chosen replica one of reference \cite{GuixGonzlez2017}, where a membrane cholesterol molecule spontaneously enters the adenosine A$_\mathrm{2A}$ receptor (A$_\mathrm{2A}$R) from the lipid phase through transmembrane 5 and 6 into the binding pocket. The time-trace shows the distance between the terminal Oxygen atom of the cholesterol and the backbone Hydrogen atom of a TM Alanine (ALA57). With this metric, several events can be appreciated and navigated to per mouse-click, throughout one $\mathrm{-long}$ trajectory: the starting point (frame 4, first arrow, top-right), the entering event (sharp drop in distance-plot at around frame 15, second arrow, top-right) and the forming of a Hydrogen bond (small drop in distance plot down to 2.91 $\AA$, frame 88, third arrow, bottom-right). The \m~session for this figure can be found at \url{http://proteinformatics.uni-leipzig.de/mdsrv_ms/fig2.html}.}
\label{NAR-fig2}
\end{figure*}

Figure \ref{NAR-fig2} shows an example of the time-trace for a selected atomic distance, which makes it possible to identify key events during the course of a trajectory.

\section*{Conclusion}

The \M web service is a simple and fast solution for sharing and interactively visualizing MD simulations in the web browser. Users can inspect MD simulations without having to install special software. The streaming approach thus eliminates the need for sophisticated hardware for use on the user side. In addition, \M offers various tools for basic analysis of MD simulations that can be used directly in the web service. \C{With the simplified installation using Docker, additional public or private instances can be easily created.}

\C{The further development of visualization tools and the requirements for a tool to easily exchange MD trajectories made further adaptations of thee \M necessary. Despite the widespread use of NGL by many web services, the switch to Mol* enabled the immediate use of some features, including superior visual quality, improved secondary structure assignment, trajectory animation video export, and fully GUI-driven session support of this visualization tool, which can be understood as an evolution of NGL and LiteMol Suite. As part of the update, we have added new features to Mol* such as support for additional coordinate and topology file formats and a Zenodo importer extension that allows easy loading of structures, trajectories, and zip archives from any Zenodo dataset. Other new features of \M include easy session creation and interactive RMSD and distance angle analysis. Finally, we have modularized the software for \M in containers to simplify installation and ensure sustainability. This improvement makes the \M user experience even more accessible and easier to use for a growing community.}

\section*{Data Availability}
We made \M accessible as open source software at 
\url{https://github.com/dwiegreffe/mdsrv}.\\
A release of the docker images is available at 
\url{https://hub.docker.com/r/dwiegreffe/mdsrv-viewer} for the viewer and 
\url{https://hub.docker.com/r/dwiegreffe/mdsrv-remote} for the remote server.\\
A hosted instance of the webservice can be accessed at 
\url{https://proteininformatics.informatik.uni-leipzig.de}.\\
A companion website with examples, FAQ and manual can be found at:
\url{https://proteininformatics.informatik.uni-leipzig.de/mdsrv/}.

\section*{ACKNOWLEDGEMENTS}

We thank all the users of our service for providing us with valuable feedback and specifically Adam Zech and Nikola Ristic for testing the latest update.

\section*{Funding}
This research was funded by the Development Bank of Saxony (SAB) under project number 100335729 (to G.S.) and the Deutsche Forschungsgemeinschaft (DFG, German Research Foundation) through CRC 1423, project number 421152132, subproject Z04 (to P.W.H.). JKST acknowledges funding by the Novo Nordisk Foundation (NNF18OC0033950, to Kresten Lindorff-Larsen).

\subsubsection*{Conflict of interest statement.} None declared.

\bibliographystyle{nar}
\bibliography{literature}

%

\end{document}